\documentclass[lettersize,journal]{IEEEtran}
\usepackage{amsmath,amsfonts}
\usepackage{algorithmic}
\usepackage{algorithm}
\usepackage{array}
\usepackage[caption=false,font=normalsize,labelfont=sf,textfont=sf]{subfig}
\usepackage{textcomp}
\usepackage{stfloats}
\usepackage{url}
\usepackage{verbatim}
\usepackage{graphicx}
\usepackage{cite}
\usepackage{booktabs}
\usepackage{multirow}
\usepackage{multicol}
\usepackage{xcolor}

\usepackage[commandnameprefix=always]{changes}

\begin{document}

\title{SDR-GAIN: A High Real-Time Occluded Pedestrian Pose Completion Method for Autonomous Driving}

\author{
Honghao Fu\textsuperscript{\rm *}, Yongli Gu\textsuperscript{\rm *}, Yidong Yan, Yilang Shen, Yiwen Wu, Libo Sun\textsuperscript{\textdagger}
\thanks{Manuscript received xx September 2023.}
\thanks{This work was supported by the Key R\&D Program of Jiangsu Province under Grant BE2023010-3. (Corresponding author: Libo Sun)}
\thanks{Honghao Fu, Yongli Gu, Yidong Yan, Yiwen Wu and Libo Sun are with the School of Instrument Science and Engineering, Southeast University, Nanjing, China.}
\thanks{Yilang Shen is affiliated with the School of Geospatial Engineering and Science, Sun Yat-sen University.}
\thanks{\textsuperscript{\rm *}These authors contributed equally.}
\thanks{\textsuperscript{\textdagger}The corresponding author.}\\
}

\markboth{Journal of \LaTeX\ Class Files,~Vol.~14, No.~8, August~2021}%
{Shell \MakeLowercase{\textit{et al.}}: A Sample Article Using IEEEtran.cls for IEEE Journals}


\maketitle

\begin{abstract}
With the advancement of vision-based autonomous driving technology, pedestrian detection have become an important component for improving traffic safety and driving system robustness. Nevertheless, in complex traffic scenarios, conventional pose estimation approaches frequently fail to accurately reconstruct occluded keypoints, primarily due to obstructions caused by vehicles, vegetation, or architectural elements. To address this issue, we propose a novel real-time occluded pedestrian pose completion framework termed Separation and Dimensionality Reduction-based Generative Adversarial Imputation Nets (SDR-GAIN). Unlike previous approaches that train visual models to distinguish occlusion patterns, SDR-GAIN aims to learn human pose directly from the numerical distribution of keypoint coordinates and interpolate missing positions. It employs a self-supervised adversarial learning paradigm to train lightweight generators with residual structures for the imputation of missing pose keypoints. Additionally, it integrates multiple pose standardization techniques to alleviate the difficulty of the learning process. Experiments conducted on the COCO and JAAD datasets demonstrate that SDR-GAIN surpasses conventional machine learning and Transformer-based missing data interpolation algorithms in accurately recovering occluded pedestrian keypoints, while simultaneously achieving microsecond-level real-time inference.
\end{abstract}


\section{Introduction}
\label{sec:intro}

\IEEEPARstart{{A}}{{utonomous}} {driving technology, as a major breakthrough in transportation, is rapidly expanding due to its convenience and efficiency~\cite{1}. Among various interactive objects, pedestrians are both the most common and among the most difficult to predict~\cite{2}, as they typically lack explicit signals to convey their intentions—unlike vehicles and infrastructure, which rely on traffic lights or physical barriers~\cite{3}. This unpredictability poses significant challenges for autonomous systems in ensuring robustness. Pedestrian detection plays a crucial role in enabling these systems to better understand pedestrian behaviors and intentions, thereby reducing accidents and improving traffic safety~\cite{4,5}. This aligns with the overarching goals of autonomous driving: to build safer and more efficient transportation systems. As a result, pedestrian detection has emerged as a key research focus in the field of autonomous driving~\cite{6}.}

{In pedestrian detection tasks, human pose estimation enhances the interpretation of pedestrian behavior for autonomous driving systems. Deep learning-based methods have gained prominence for their improved accuracy and autonomy. For example, regression-based methods predict keypoint coordinates directly from images in an end-to-end manner~\cite{7,8,9}, but they often suffer from high computational costs and limited spatial accuracy. Heatmap-based methods generate 2D heatmaps by placing Gaussian kernels at each keypoint~\cite{10}, better preserving spatial information~\cite{11,12}, though accuracy declines at lower resolutions. Top-down approaches individually localize pedestrians and estimate pose for each region~\cite{13,14}, but struggle with real-time performance in multi-person scenarios due to increasing iterations. In contrast, bottom-up methods detect all keypoints and group them into individual poses through post-processing~\cite{15,16,17}, significantly improving computational efficiency. In real-world autonomous driving environments, characterized by high vehicle speeds and dynamic complexity, pose estimation systems must support multi-person detection with both high accuracy and low latency~\cite{18}. Therefore, bottom-up methods are considered a promising solution due to their superior real-time performance and efficiency in handling crowded situations.}

{With the advancement of deep learning, bottom-up methods have achieved sufficient accuracy for real-world applications while maintaining real-time performance~\cite{20,21}. However, pedestrian occlusion by vehicles, vegetation, or buildings in traffic scenes remains a significant challenge, making the improvement of keypoint detection under occlusion essential for robust pose estimation. Common solutions involve training pedestrian detection or pose estimation models to handle specific occlusion patterns~\cite{46,48}, or classifying occlusion types~\cite{47} to enhance robustness under various occlusion conditions. However, these vision-based approaches for occluded pedestrian detection often suffer from inference latency. Thus, an open problem remains in real-time–critical applications such as autonomous driving:} 

\textit{{- How to accurate pose estimation for occluded pedestrian while maintaining high real-time performance?}}


{To address this, we propose a novel occlusion-aware human pose completion method, named Separation and Dimensionality Reduction-based Generative Adversarial Imputation Networks (SDR-GAIN). Unlike previous approaches that train visual models to distinguish occlusion patterns~\cite{46,47,48}, SDR-GAIN aims to learn human pose directly from the numerical distribution of keypoint coordinates and interpolate missing positions, thereby enabling a more lightweight model architecture. Specifically, SDR-GAIN adopts a self-supervised generative adversarial framework with a mask and hint mechanism, which leverages the intrinsic structural information of the data without requiring additional annotations. By employing a lightweight network architecture with residual structures, our method ensures high pose estimation performance while significantly reducing model complexity and enhancing real-time capability, making it particularly suitable for practical autonomous driving applications. Furthermore, to mitigate the challenges of adversarial learning, we enhance the pose feature representation and standardize the pose data distribution through processes such as separation, rotation, and dimensionality reduction. Experimental results demonstrate that our proposed method outperforms both traditional machine learning and emerging Transformer-based deep learning interpolation algorithms in the task of missing pose completion. Our key contributions are summarized as follows:}
\begin{itemize}
    \item {We propose SDR-GAIN, a lightweight self-supervised method that learns human pose directly from the numerical distribution of keypoint coordinates, rather than training visual models to distinguish occlusion patterns, thereby balancing real-time performance and accuracy.}
       
    \item {SDR-GAIN uses pose data standardization methods and employs a multi-generator strategy to separately learn head and torso poses with distinct distributions, reducing the learning difficulty and improving the accuracy of occluded keypoint imputation.}

    \item {Experiments on multiple datasets indicate that SDR-GAIN achieves at least a 47.4\% reduction in RMSE compared to other baselines. With microsecond-level inference time, it is highly computationally efficient for real-time applications in autonomous driving.}
       
\end{itemize}


\section{Related Works}
\label{sec:rw}

\subsection{Pedestrian pose estimation and detection}
To accurately infer the poses of occluded pedestrians, precise keypoint estimation~\cite{cai1,cai2,cai3} is crucial. In traffic scenes with multiple pedestrians, top-down methods~\cite{30,31} often face challenges in real-time performance. In contrast, bottom-up methods~\cite{32,33} present a promising alternative. A key advancement in this area is OpenPose~\cite{34}, which employs the Part Affinity Fields (PAFs) to associate body parts with individuals while encoding the global context, supported by a bottom-up parsing step to ensure both high accuracy and real-time processing. With ongoing improvements in speed and accuracy~\cite{35}, bottom-up methods show significant potential for real-time applications, further fostering the adoption of posture-based approaches in pedestrian detection tasks~\cite{41,cai4}, such as behavior recognition~\cite{36,37} and action prediction~\cite{38,39,42,43}. Marginean et al.\cite{44} proposed posture-based features and recursive architectures to classify short sequences of pedestrian behaviors. Zhang et al.\cite{40} generated skeleton keypoints from CCTV footage and employed machine learning models to predict pedestrian intentions at traffic lights. However, despite the high accuracy of posture-based methods, the reliance on predefined keypoints presents a challenge, as occlusions may result in missing keypoints, reducing the robustness of these detection systems.

\subsection{Pose estimation of occluded pedestrians}
Since pedestrian detection methods often experience performance degradation due to inevitable occlusion~\cite{45,49}, minimizing the negative impact of occlusion is critical for improving detection robustness. Ouyang et al.~\cite{46} proposed a deep learning model that learns the visibility relationships between overlapping human poses, offering a versatile post-processing approach for pose estimation results. Mathias et al.~\cite{47} proposed to train classifiers for specific occlusion scenarios by reusing computations across different training stages. However, the use of multiple classifiers for distinct occlusion conditions compromises real-time performance due to high computational costs. Zhang et al.~\cite{48} employed cross-channel attention mechanisms to capture various occlusion patterns within a single model, enhancing a FasterRCNN detector with attention networks for detecting occluded pedestrians. However, existing vision-based occluded pedestrian detection methods exhibit significant inference delay, presenting challenges for high real-time tasks such as autonomous driving.

\subsection{Generative models}
Recent advancements in generative models have introduced potential solutions for addressing missing keypoints in pose estimation. Goodfellow et al.\cite{50} proposed generative adversarial networks (GANs), which utilize adversarial learning to drive data generation. This framework involves training a generator to capture data distributions and a discriminator to distinguish between real data and generated samples. Yoon et al.\cite{51} extended the GAN framework by introducing generative adversarial imputation network (GAIN), which improves missing data generation by providing the discriminator with hint mask that highlight the missing parts of the ground truth sample. Ho et al.\cite{52} drew inspiration from non-equilibrium thermodynamics to propose a 2D generation method using diffusion probability models, trained with a weighted variation constraint linked to Langevin dynamics. The success of generative models in handling various types of missing data\cite{53,54,55,sup1} demonstrates their efficacy. Building on these advancements, we aim to implement an accurate, real-time occlusion-resistant pose interpolation method based on generative models.

\section{{Methodology of SDR-GAIN}}
\label{sec:method}
{
In this section, we present the methodology of our proposed SDR-GAIN. In the framework, pedestrian pose keypoints are first estimated from input images using deep learning-based computer vision algorithms. Then, we improve the pose feature representation and standardize the data distribution through processes including separation, rotation, and dimensionality reduction. Subsequently, GANs trained with a masked learning strategy are employed to impute missing keypoints resulting from partial occlusion. The following subsections specifically describe the components of SDR-GAIN.}

{\subsection{Overview}}
\begin{figure*}[t]
  \centering
  \includegraphics[width=0.9\linewidth]{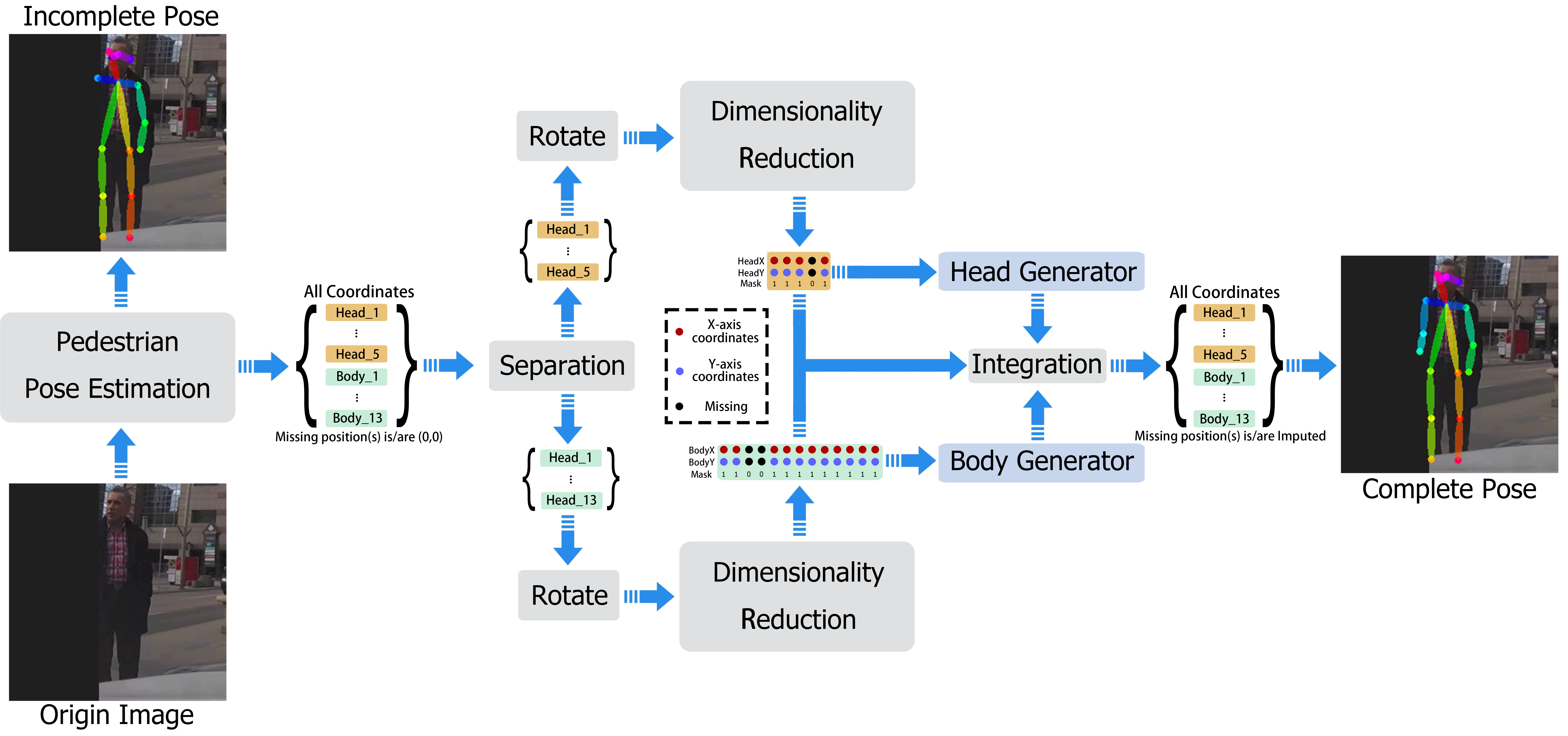}
  \caption{The framework of SDR-GAIN. It starts with initial pose estimation via OpenPose, followed by separation, rotation and dimensionality reduction methods for pose data standardization. Then, two generators are used to predict the missing keypoints for the head and torso. The generator's output is then reverse-processed to restore the complete pedestrian pose.}
  \label{fm}
\end{figure*}

Fig.~\ref{fm} illustrates the framework of SDR-GAIN, which completes missing poses of occluded pedestrians through keypoint estimation and generative models. It begins with initial pose estimation using OpenPose on an image or video frame, followed by reversible separation and rotation of missing head and torso keypoints based on reference points (e.g., ears and shoulders) to standardize the feature distribution. The keypoints are then projected into one-dimensional distributions, normalized, and merged to enhance uniformity. Next, we use mask vectors to hint the missing keypoints. We train two GAN-based generative models, one for the head and one for the torso, which accept the one-dimensional projections of the keypoints with missing data as input and use the hint mask vectors to indicate the missing positions. Finally, the generator’s output is reverse-processed to restore the keypoint coordinates and complete the pedestrian pose.

\begin{figure}[t]
  \centering
  \includegraphics[width=\linewidth]{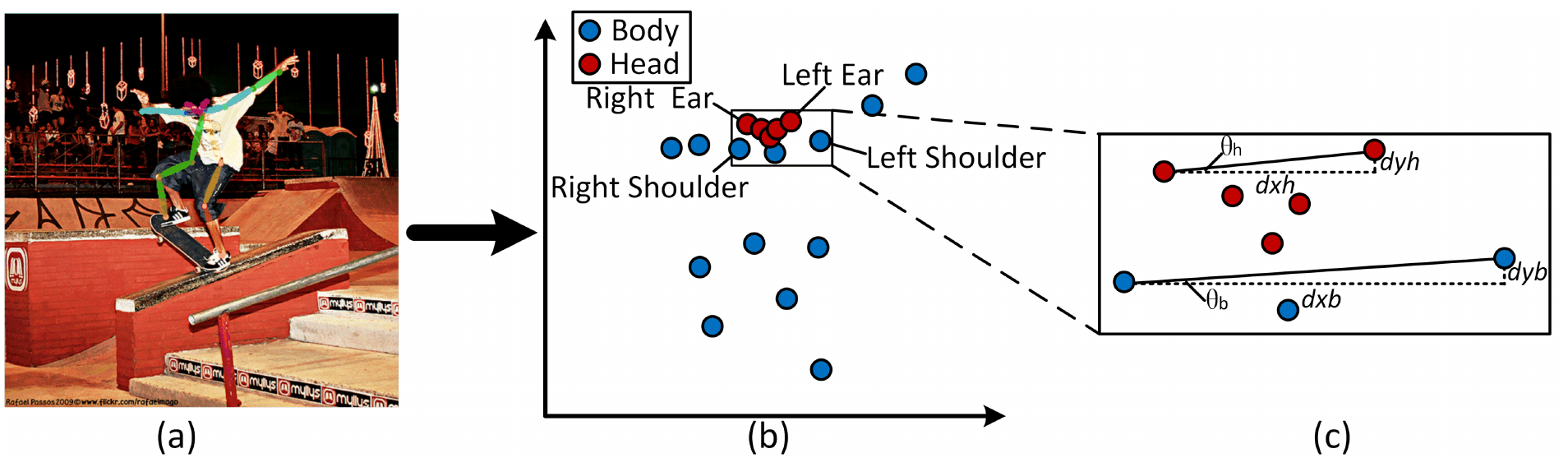}
  \caption{Separation and calculation of rotation angle. (a) Image after pose estimation; (b) Keypoints map, where blue and red dots represent keypoints of the torso and head, respectively; (c) Calculation of rotation angle with reference to the right and left ears, and the right and left shoulders.}
  \label{rot}
\end{figure}
{\subsection{Pose estimation and standardization}}
\noindent\textbf{Pose estimation.} Pedestrian pose estimation aims to extract skeletal keypoints from pedestrian postures in each frame of an image or video, represented as a two-dimensional distribution. OpenPose~\cite{34} is a widely adopted method for this task~\cite{57,58}, particularly in autonomous driving research~\cite{6}. We use the classic 18-point model proposed by OpenPose, based on the COCO 17-point model~\cite{64}, as the standard for pose estimation and completion. The model includes five head keypoints and thirteen body keypoints. Fig.~\ref{rot}(a) shows the results of pose estimation using OpenPose.

\noindent\textbf{Separation and rotation.} The dispersion of head keypoints relative to the torso, as illustrated in Fig.~\ref{rot}(a), demonstrates a notable difference in their distribution. This suggests that training a single model on both head and torso keypoints may impede the learning of distinct features. To mitigate this issue, the five head keypoints and thirteen body keypoints predefined by OpenPose are extracted and utilized separately to train two distinct GANs. Each model generates specific outputs for the missing head or torso keypoints, as shown in Fig.~\ref{rot}(b). Due to the variability in human posture tilt angles across images, which raw coordinate data may not fully capture, rotating the coordinates to a consistent angle prior to input into the GAN helps standardize the keypoint distributions, thus facilitating more effective feature learning.

As a rotational reference, we select the symmetrical and most distantly spaced keypoint pair (left and right ears) for the head coordinates, and the only symmetrical keypoint pair (left and right shoulders) for the torso. Specifically, the set of head keypoints are denoted by $H \in \mathbb{N}^2$. As shown in Fig.~\ref{rot}(c), the rotation center is chosen as the right ear. The rotation angle $\theta_h$ is defined as the acute angle between the line connecting the right ear and the left ear and the $x$-axis:
\begin{equation}
\theta_h = \arctan \left( \frac{d y_h}{d x_h} \right)
\end{equation}
Where $d x_h$ and $d y_h$ represent the distances between these two points along the $x$ and $y$ axes, respectively. Let $\mathbf{B} \in \mathbb{N}^2$ represent the set of keypoints for torso. The rotation center is the right shoulder. The rotation angle $\theta_b$ is the acute angle formed by the line connecting the right and left shoulder points and the $x$-axis:
\begin{equation}
\theta_b = \arctan \left( \frac{d y_b}{d x_b} \right)
\end{equation}
Where $d x_b$ and $d y_b$ denote the distances between these two points along the $x$ and $y$ axes, respectively.

\noindent\textbf{Dimensionality reduction.} We performed dimensionality reduction on the posture coordinate data by projecting each element of \( H \) and \( B \) onto the \( x \)- and \( y \)-axes and normalizing them. Specifically, Fig.~3(a) and (b) shows the dimensionality reduction for the torso and head keypoints, respectively. The red lines in the figure indicate that the torso and head were rotated to a specified position before the dimensionality reduction process. The training set for the coordinate data is denoted as \( S \in \mathbb{R}^3 \), and we define the vector projections of the reduced torso and head keypoints in the \( x \)- and \( y \)-directions corresponding to the \( d \)-th image's coordinate data \( S_d \) as \( B_{xd}, B_{yd}, H_{xd}, H_{yd} \in \mathbb{R} \), respectively. Furthermore, we define \( S_d = \{B_{xd}, B_{yd}, H_{xd}, H_{yd}\} \in \mathbb{R}^2 \), and let \( S_{d,i,k} \) denote the \( k \)-th element of the \( i \)-th vector \( S_{d,i} \) in \( S_d \). We then define \( NS_d \) as the set of all elements in \( S_d \) after normalization and normalize all elements in \( S_{d,i} \) based on the following formula:
\begin{equation}
NS_{d,i,k} = \frac{S_{d,i,k} - \min(S_{d,i,k})}{\max(S_{d,i,k}) - \min(S_{d,i,k})}   
\end{equation}
Where $NS_{d,i,k}$ denotes the $k$-th component of the $i$-th vector $NS_{d,i} \in [0,1]^k$ in the set $NS_d$. The goal of normalization is to map data with diverse distributions to the interval $[0,1]$, thereby standardizing the probability distribution and minimizing the influence of anomalous samples. 
\begin{figure}[t]
  \centering
  \includegraphics[width=\linewidth]{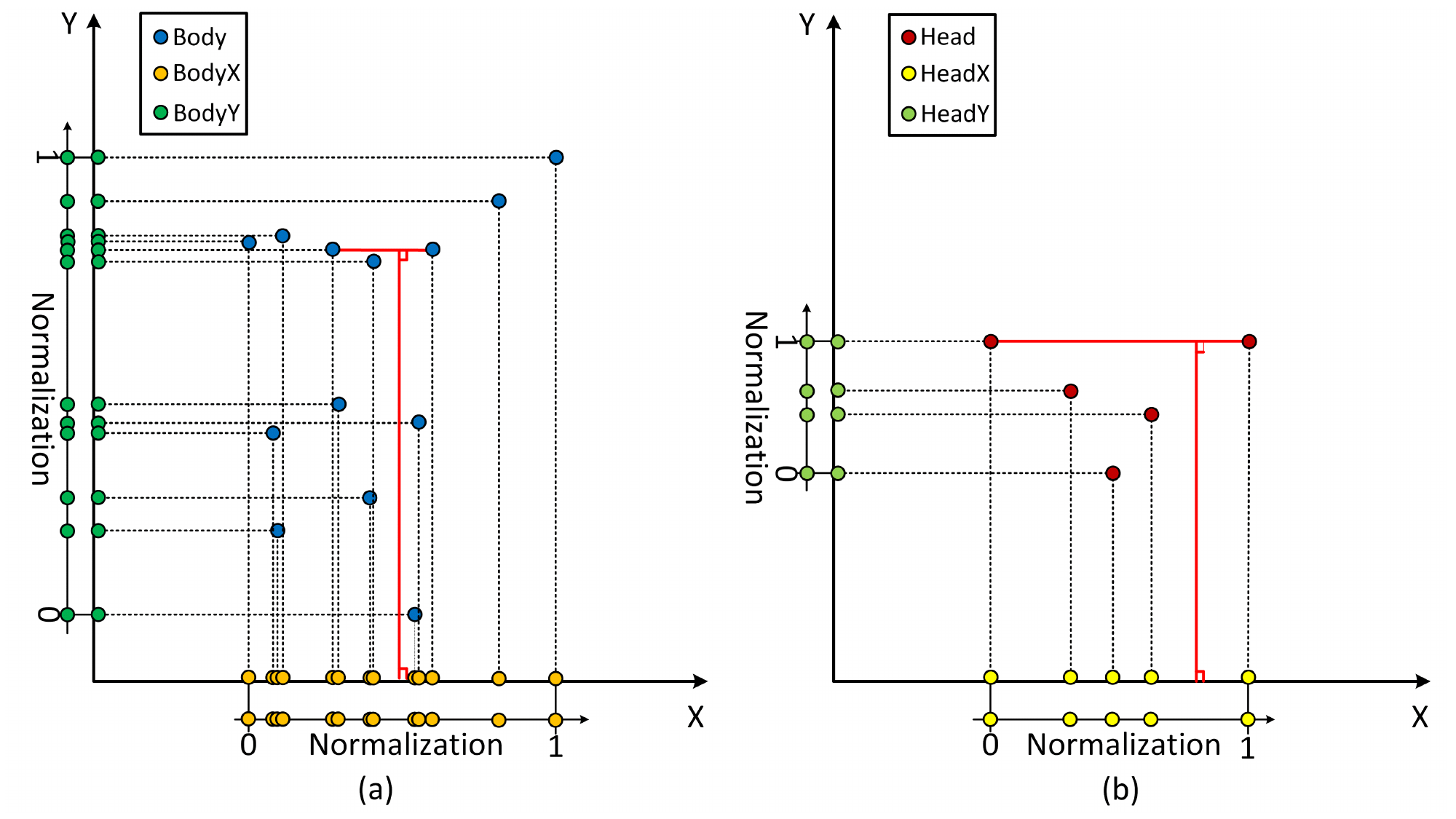}
  \caption{Dimensionality reduction and normalization process: (a) Applied to the rotated torso keypoints, with the blue point representing the original 2D coordinates and the orange and dark green points representing the reduced coordinates along the $x$ and $y$ axes, respectively; (b) Applied to the rotated head keypoints, with the red point representing the original 2D coordinates and the yellow and light green points representing the reduced coordinates along the $x$ and $y$ axes, respectively.}
  \label{dr}
\end{figure}

\subsection{Training}
\begin{figure*}[t]
  \centering
  \includegraphics[width=\linewidth]{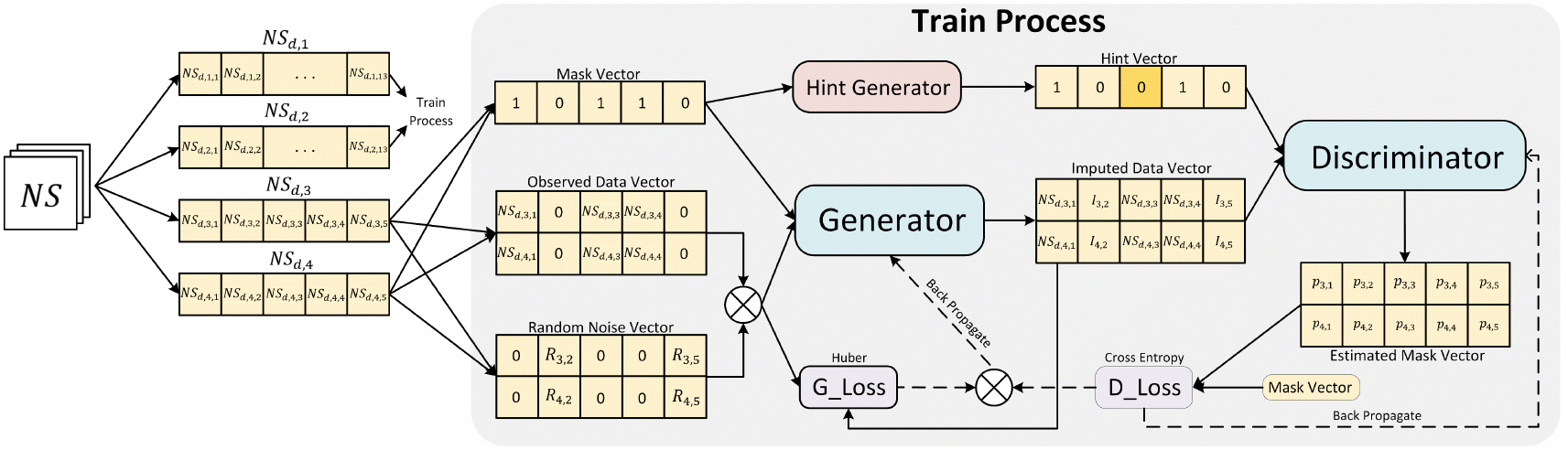}
  \caption{The training process of generators and discriminators.}
  \label{train}
\end{figure*}

\noindent\textbf{Preliminary.} GAN is a widely used deep learning framework consisting of two networks: the generator and the discriminator~\cite{50}. The generator learns to model the input data distribution and generates synthetic data that closely resembles the training data, while the discriminator distinguishes between real and generated data. These networks engage in an adversarial process, with the generator attempting to deceive the discriminator, and the discriminator working to correctly classify real and fake data. Through iterative optimization, the generator improves its output, producing more realistic data. GAIN~\cite{51}, a GAN extension, has shown success in missing data imputation~\cite{61,62}. However, human posture data presents unique challenges due to its two-dimensional spatial distribution and interdependent keypoints, which are not simply governed by a data distribution. Extracting these spatial relationships adds complexity. To address this, we adapt the GAIN framework and enhance the neural network by introducing residual structures~\cite{63}. These residual connections allow certain layers to bypass others, facilitating deeper learning and mitigating issues like vanishing gradients and overfitting. Regularization is also applied to ensure generalizability. The training process for both the generator and discriminator is outlined in Fig.~\ref{train}.

\begin{figure*}[t]
  \centering
  \includegraphics[width=0.7\linewidth]{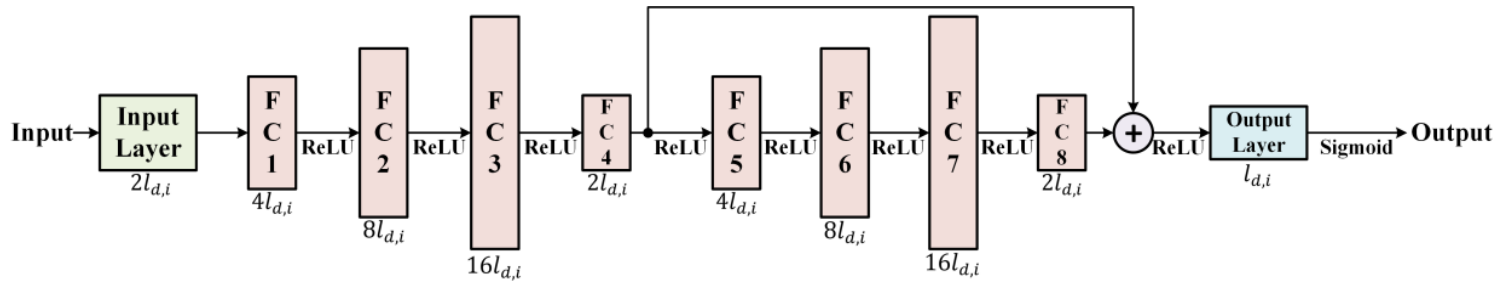}
  \caption{Network structure.}
  \label{structure}
\end{figure*}

\noindent\textbf{Masked learning.} We construct data with missing values using the mask vector and employ the hint vector to indicate the missing positions during training. Firstly, the posture vectors obtained via separation and dimensionality reduction are treated as distinct datasets, following the same procedure for each vector. Let $p_m \in (0,1)$ denote the data loss rate. For a vector $NS_{d,i}$ of length $l_{d,i}$, the corresponding mask vector is $M_{d,i} \in {0,1}^{l_{d,i}}$. Specifically, the $k$-th element of the mask $M_{d,i,k}$ is defined as:
\begin{equation}
M_{d,i,k} = 
\begin{cases}
1, & \text{if } A_k > p_m \\
0, & \text{otherwise}
\end{cases}    
\end{equation}
Where $A_k$ denotes a randomly generated vector of length $l_{d,i}$, where the $k$-th element belongs to $A \in [0,1]^{l_{d,i}}$. Moreover, each element of $M_{d,i}$ encodes specific pose keypoint information: an element with value 1 indicates the keypoint is present, while a value of 0 indicates the keypoint is missing. Let the masked observed data vector be denoted as $OS_{d,i}$, with the data at its $k$-th position given by:
\begin{equation}
OS_{d,i,k} = 
\begin{cases}
NS_{d,i,k}, & \text{if } M_{d,i,k} = 1 \\
0, & \text{otherwise}
\end{cases}
\end{equation}
Let $NR \in [0,1]^{l_{d,i}}$ be a random noise vector of length $l_{d,i}$. The corresponding interpolated random noise vector for $NS_{d,i}$ is denoted as $R_{d,i}$, defined as follows:
\begin{equation}
R_{d,i} = (1 - M_{d,i}) \cdot NR
\end{equation}

\noindent\textbf{Generator.} To improve the learning of pose features, we introduce residual structures and deepen the generator and discriminator layers. The generator, denoted as $G$, includes a hidden layer with eight ReLU layers and residual modules, as shown in Fig.~\ref{structure}. A residual connection is applied after the eighth ReLU layer, where the outputs of the fourth and eighth layers are summed and fed into a Sigmoid output layer. Let $G$ be a mathematical function. Assuming $IS_{d,i} = OS_{d,i} + R_{d,i}$ and $M_{d,i}$ are the inputs to $G$, the generated value of $G$ is denoted as $IG_{(d,i)} =G(IS_{d,i},M_{d,i})$, where $IG_{(d,i)} \in [0,1]^{l_{(d,i)}}$. In the next step, the non-missing data is superimposed onto the corresponding positions in the matrix $IG_{d,i}$, forming a new vector $I_{d,i}$, where the $k$-th data point is denoted as $I_{d,i,k}$. This process is expressed as follows:
\begin{equation}
I_{(d,i,k)} =
\begin{cases}
NS_{d,i,k}, & \text{if } M_{d,i,k} = 1 \\
IG_{d,i,k}, & \text{otherwise}
\end{cases}    
\end{equation}

\noindent\textbf{Discriminator.} Let $D$ be the discriminator, with a structure identical to that of the generator. Assuming $D$ is a mathematical function, with inputs $I_{d,i}$ and $H_{d,i} \in {0,1}^{l_{d,i}}$, where $H_{d,i}$ is a hint vector. The output $E_{d,i} \in [0,1]^{l_{d,i}}$ of $D$ represents the confidence that the generated data is real. The process is expressed as follows:
\begin{equation}
E_{d,i} = D(I_{d,i}, H_{d,i})
\end{equation}
The generation of the hint vector $H_{d,i}$ follows a process similar to that of $M_{d,i}$. Let $C_k$ denote the $k$-th data in a random vector $C \in [0,1]^{l_{d,i}}$ of length $l_{d,i}$. Let $p_h \in (0,1)$ represent the hint rate, and the $k$-th data point $H_{d,i,k}$ of $H_{d,i}$ is given by:
\begin{equation}
H_{d,i,k} =
\begin{cases}
M_{d,i,k}, & \text{if } C_k > 1 - p_h \\
0, & \text{otherwise}
\end{cases}
\end{equation}

\noindent\textbf{Adversarial optimization.} To train the discriminator $D$ and the generator $G$ via error backpropagation, loss functions must be defined for each. Let $L_{(D,i)}(a,b)$ denote the loss function for $D$. To quantify the difference between the generated and real data distributions, cross-entropy is used as the loss function for $D$, expressed as:
\begin{equation}
L_{\text{CE}}(a, b) = - \left( b \cdot \log(a) + (1 - b) \cdot \log(1 - a) \right)
\end{equation}
Where $a = E_{d,i,k}$ and $b = M_{d,i,k}$ are the $k$-th elements of $E_{d,i}$ and $M_{d,i}$, respectively. Averaging the $L_{D,i,k}$ values across all bits of a given input sample yields the loss function $L_{D,i}$ for $D$ with respect to the current $S_{d,i}$ input sample. The index range is defined as $i \in {1, 2, 3, 4}$, consistent with prior definitions.
\begin{equation}
L_{D,i}\left(E_{d,i}, M_{d,i}\right) = \frac{1}{l_{d,i}} \sum_{k=1}^{l_{d,i}} L_{CE}\left(E_{d,i,k}, M_{d,i,k}\right)
\end{equation}
The loss function plays a crucial role in training $D$. When $M_{d,i,k} = 1$, $E_{d,i,k}$ represents the confidence score of $D$ for real data. The objective is to maximize $E_{d,i,k}$ towards 1, minimizing $-\log E_{d,i,k}$. Conversely, when $M_{d,i,k} = 0$, $E_{d,i,k}$ represents $D$'s confidence for generated data. Here, the goal is to minimize $E_{d,i,k}$ towards 0, which minimizes $-\log(1 - E_{d,i,k})$. Through this iterative process, $D$ is effectively trained. Let the length of the coordinate dataset $S$ be denoted as $l_s$. The training procedure for $D$ is summarized as follows:
\begin{equation}
\min_D \sum_{d=1}^{l_s} L_{D,i} \left( E_{d,i}, M_{d,i} \right)
\end{equation}
To train the discriminator $D$, updates are made through stochastic gradient descent (SGD).
\begin{equation}
\nabla_D \sum_{d=1}^{l_s} L_{D,i} \left( E_{d,i}, M_{d,i} \right)
\end{equation}
\begin{table*}[t]
  \caption{{Performance comparison between SDR-GAIN and other imputation algorithms. Bold entries indicate the best results.}}
  \centering
  \footnotesize
  \setlength{\tabcolsep}{2pt}
  \begin{tabular}{l|c|cccccccc}
    \toprule
    \multirow{2}{*}{Dataset} & \multirow{2}{*}{Metrics} & \multicolumn{8}{c}{Interpolation Algorithms}\\ \cmidrule(l){3-10}
    & &\textbf{SDR-GAIN}&GAIN&Transformer&Reformer&Pyraformer&Corssformer&k-NN&MissForest \\
    \midrule
    \multirow{2}{*}{COCO}&RMSE &\textbf{0.0225}&0.0768&0.0506&0.0463&0.0494&0.0428&0.3406&0.3392\\
    & Time/s &$4.580 \times 10^{-4}$&$1.128 \times 10^{-4}$&$3.476 \times 10^{-3}$&$7.412 \times 10^{-3}$&$6.238 \times 10^{-3}$&$1.060 \times 10^{-2}$&$2.131 \times 10^{-4}$&$3.560 \times 10^{-3}$\\
    \midrule
    \multirow{2}{*}{JAAD}&RMSE &\textbf{0.0117}&0.0672&0.0429&0.0402&0.0426&0.0435&0.6171&0.5936\\
    & Time/s &$4.087 \times 10^{-4}$&$1.077 \times 10^{-4}$&$3.273 \times 10^{-3}$&$6.959 \times 10^{-3}$&$6.113 \times 10^{-3}$&$9.028 \times 10^{-3}$&$8.336 \times 10^{-5}$&$7.824 \times 10^{-3}$\\
    \bottomrule
  \end{tabular}
  \label{tab:cp1}
\end{table*}
The latest update of $D$ is then used to train $G$. Since $G$ generates values not only for missing data points but also for non-missing ones through neural network processing, solely evaluating the discrepancy at missing data points is insufficient for $G$ to learn the distribution of real data. Therefore, it is essential to assess the similarity between generated values at non-missing data points and their original counterparts. Based on the most recent $D$ update, the loss function $L_M$ is defined to evaluate the quality of generated values at missing data points as follows:
\begin{equation}
L_M (E_{d,i}, M_{d,i}) = -\frac{1}{l_{d,i}} \sum_{k=1}^{l_{d,i}} (1 - M_{d,i,k}) \log(E_{d,i,k})
\end{equation}
During model training with gradient descent, mean squared error (MSE) typically offers better accuracy than mean absolute error (MAE). However, MAE is more robust to outliers than MSE. Thus, we use the Huber loss function to evaluate the similarity between generated and true values at non-missing positions. To accurately reflect the generator’s performance in learning the true data distribution, the loss is computed using data that is not marked as missing. This requires introducing a mask variable $m$. Therefore, we define the Huber loss used by the generator $G$ as $L_{\text{Huber}}(y, f(x), m)$:
\begin{equation}
{
\begin{aligned}
&L_{\text{Huber}}(y, \hat{y}, m) = 
\begin{cases}
0.5m^2(y - \hat{y})^2, & \text{if } |y - \hat{y}| \leq \delta, \\
\delta |m(y - \hat{y})| - 0.5 \delta^2, & \text{otherwise}.
\end{cases}
\end{aligned}
}
\end{equation}
Where $y = IS_{d,i}$, $\hat{y} = IG_{d,i}$, and $m = M_{d,i}$, with $\delta$ as a hyperparameter termed Huber threshold. When the prediction error is less than or equal to $\delta$, the Huber loss approximates MAE; when it exceeds $\delta$, it approximates MSE. To prevent overfitting, L1 regularization is applied to the neural network. The penalty $p$ is the sum of the absolute values of all weights, with $\lambda$ as the regularization weight. The resulting loss function $L_{(G,i)}$ for $G$ with respect to the current input sample $S_{(d,i)}$ is then derived:
\begin{equation}
L_{G,i} = \alpha L_{\text{Huber}}(IS_{d,i}, IG_{d,i}, M_{d,i}) + L_M(E_{d,i}, M_{d,i}) + \lambda p
\end{equation}
Where $\alpha$ is loss weight. The goal of this loss function is to optimize the generator $G$ to produce data in the missing regions that are convincingly classified as real by the discriminator $D$ when $M_{d,i,k}=0$. This requires minimizing $L_{M,i}$ and maximizing $E_{d,i,k}$ towards 1. When $M_{d,i,k}=1$, the objective is to minimize $L_{\text{Huber}}$, ensuring that the values generated by $G$ in the non-missing areas closely match the actual values. The hyperparameter $\alpha$ controls the influence of $L_{\text{Huber}}$ on $L_{G,i}$, thereby regulating $G$'s performance optimization. The training procedure for $G$ is summarized as follows:
\begin{equation}
\min_G \sum_{d=1}^{l_s} L_{G,i} \left( IS_{d,i}, IG_{d,i}, M_{d,i}, E_{d,i} \right)
\end{equation}
To train the $G$, updates are made through SGD:
\begin{equation}
\nabla_G \sum_{d=1}^{l_s} L_{G,i} \left( IS_{d,i}, IG_{d,i}, M_{d,i}, E_{d,i} \right)
\end{equation}

\section{Experiments and evaluations}
\label{sec:exp}

\subsection{Datasets and Implementation}
\noindent\textbf{Datasets.} We train and evaluate our models using the COCO and JAAD datasets. COCO~\cite{64} is a large-scale dataset with keypoint annotations for over 100K human bodies, and JAAD~\cite{65} captures the behavioral variability of traffic participants. These two datasets are extensively used in pedestrian pose estimation~\cite{42,66}. We conduct comparative experiments on both datasets, and ablation study on COCO.

\noindent\textbf{Implementation.} The training platform consists of a 3060Ti GPU, an i7-12700F CPU, and 32GB of RAM. Unless otherwise stated, the testing platform matches the training setup, and the hyperparameters used in the method are set to default conditions as follows: data missing rate $p_m = 0.2$, hint rate $p_h = 0.9$, Huber weight $\delta = 0.6$, loss weight $\alpha = 10$ and regularization weight$\lambda = 0.001$. {We we randomly split datasets into training and testing sets in 8:2 and train the models using a batch size of 128 for 5000 epochs. Moreover, we summarize the definitions of the variables in Section~\ref{sec:method} in Table A of the supplementary material.}


\subsection{Comparison against other methods}
\noindent\textbf{{Quantitative comparison.}} Table~\ref{tab:cp1} compares the proposed SDR-GAIN with other imputation methods in terms of accuracy (RMSE) and real-time capability (Time/s) for human pose completion. The reported time indicates the average duration required to complete one human pose. The baselines include the original GAIN framework~\cite{51}, Transformer~\cite{Transfor} and its variants (Reformer~\cite{Refor}, Pyraformer~\cite{Pyra}, Crossformer~\cite{Cross}), as well as traditional machine learning methods such as k-NN~\cite{69} and MissForest~\cite{70}. The results show that SDR-GAIN achieves higher accuracy than both deep learning and classical machine learning baselines. Notably, it maintains microsecond-level real-time performance comparable to efficiency-focused methods (GAIN, k-NN), while delivering performance improvements by an order of magnitude over precision-focused Transformer-based approaches. Furthermore, SDR-GAIN exhibits significantly better performance on the JAAD dataset, highlighting its strong potential for applications in traffic-related scenarios. In addition, we provide the parameter size and computational complexity of SDR-GAIN and the baseline models in Table B of the supplementary material.
\begin{table}[t]
  \caption{Learning rate decay at which epoch.}
  \centering
  \small
  \begin{tabular}{clcl}
    \toprule
    Number     &  Keypoint   &  Number     &  Keypoint \\
    \midrule
    0&Nose&9&Right Knee \\
    1&Neck&10& Right Ankle\\
    2&Right Shoulder&11&Left Hip \\
    3&Right Elbow&12&Left Knee  \\
    4&Right Wrist&13& Left Ankle\\
    5&Left Shoulder&14&Right Eye\\
    6&Left Elbow&15&Left Eye \\
    7&Left Wrist&16&Right Ear \\
    8&Right Hip&17&Left Ear \\
    \bottomrule
  \end{tabular}
  \label{tab:keypoints}
\end{table}

\begin{figure}[t]
  \centering
\includegraphics[width=0.9\linewidth]{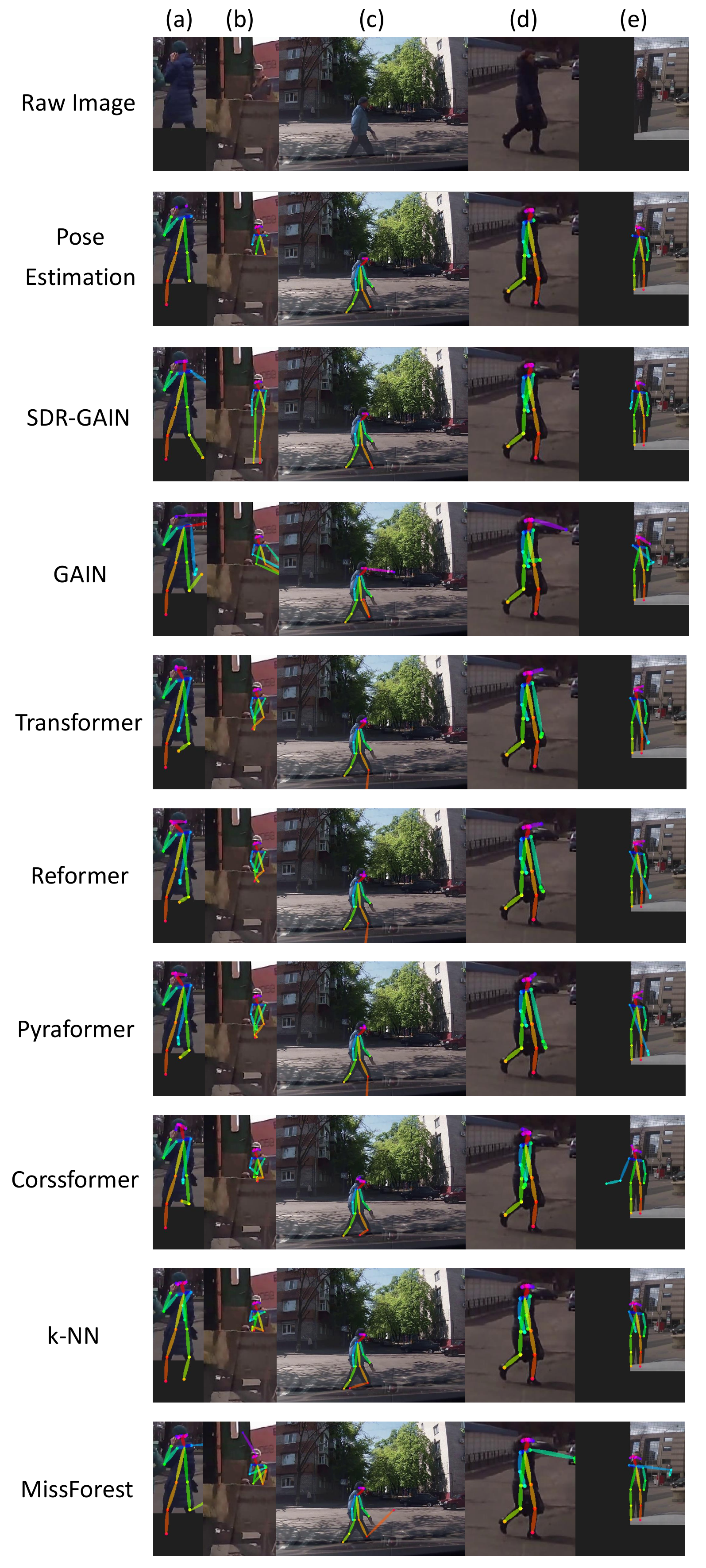}
  \caption{{Qualitative comparison between SDR-GAIN and other baselines.}}
  \label{fig:cp1}
  \vspace{-0.1 in}
\end{figure}

\noindent\textbf{{Qualitative comparison.}} As shown in Fig.~\ref{fig:cp1}, we qualitatively compare the performance of various imputation methods for human pose completion on the JAAD dataset. Table~\ref{tab:keypoints} provides the mapping between keypoint indices and corresponding body parts. In Fig.~\ref{fig:cp1}(a), foot and hand keypoints are missing due to image boundary limitations and self-occlusion; keypoints 0, 3, 4, 10, 14, and 15 are completed to address this. In Fig.~\ref{fig:cp1}(b), occlusion by a large object leads to significant loss of torso keypoints, with keypoints 4, 6, 7, 9, 10, 12, and 13 being restored. Fig.~\ref{fig:cp1}(c) shows missing foot keypoints due to occlusion by a car hood, addressed by completing keypoints 13, 15, and 17. In Fig.~\ref{fig:cp1}(d), arm keypoints are lost due to self-occlusion; keypoints 6, 7, and 17are recovered. Lastly, in Fig.~\ref{fig:cp1}(e), arm keypoints are missing due to image boundary issues, with keypoints 3, 4, and 16 completed. The dark gray areas in Fig.~\ref{fig:cp1} indicate regions beyond the image boundary. The qualitative results demonstrate that SDR-GAIN consistently outperforms other methods, highlighting its ability to accurately learn the true distribution of human poses from coordinate data.

\subsection{Ablation Study}
\begin{table*}[t]
  \caption{Ablation study on the performance of generators with different loss functions and structures under separation and dimensionality reduction strategies (RMSE). Bold entries indicate the best results.'w/o', 'S', 'DR', 'MSE', 'Huber' and 'Res' indicate 'without', separation, dimensionality reduction, MSE loss, Huber loss and residual connection, respectively.}
  \centering
  \footnotesize
  \setlength{\tabcolsep}{3pt}
  \begin{tabular}{l|cccccccccccc}
    \toprule
    \multirow{2}{*}{Loss \& Structure} & \multicolumn{6}{c}{SDR-GAIN} & \multicolumn{2}{c}{w/o DR} &  \multicolumn{3}{c}{w/o S} &  w/o S \& DR \\ \cmidrule(l){2-7} \cmidrule(l){8-9} \cmidrule(l){10-12} \cmidrule(l){13-13}
    &HeadD &HeadX &HeadY &BodyD  &BodyX &BodyY &Head &Body &BothD &BothX &BothY &Both \\
    \midrule
    MSE \& Res& 0.4527&0.1362& 0.1834&0.3470& 0.2281&0.2485  &0.2233 &\textbf{0.1383} &0.3537 &\textbf{0.1969} &\textbf{0.1972} &\textbf{0.2113} \\
    Huber \& Res& 0.4391&0.1441 &0.1664 &\textbf{0.3341}  &\textbf{0.2173} &\textbf{0.2309} &\textbf{0.1716}  &0.2046 &\textbf{0.3257} &0.2265 &0.2306 &0.2260 \\
    MSE w/o Res&0.4351 &0.1571 &0.1797 &0.3726 &0.2375 &0.2578 &0.3576 &0.2414 &0.3434 &0.2122 &0.2377 &0.2131 \\
    Huber w/o Res&\textbf{0.4022} &\textbf{0.1318} &\textbf{0.1648} &0.3559 &0.2331 &0.2505 &0.1888 &0.2768 &0.3372 &0.2321 &0.2323 &0.2271 \\
    \bottomrule
  \end{tabular}
  \label{tab:ab1}
\end{table*}
\begin{table}[t]
  \caption{Ablation study on the seperation, dimensionality reduction and multi-generator strategies (RMSE). Bold entries indicate the best results, while '-' are not applicable.}
  \centering
  \footnotesize
  \setlength{\tabcolsep}{2.2pt}
  \begin{tabular}{l|cccccc}
    \toprule
    \multirow{2}{*}{Loss \& Structure} & \multicolumn{2}{c}{SDR-GAIN} & w/o DR &  \multicolumn{2}{c}{w/o S} &  w/o S \& DR \\ \cmidrule(l){2-3} \cmidrule(l){4-4} \cmidrule(l){5-6} \cmidrule(l){7-7}
    &Dev&No Dev &- &Dev  &No Dev &- \\
    \midrule
    MSE \& Res& 0.0401&\textbf{0.0242} &0.0839 &0.0489 &0.0454 &0.0929 \\
    Huber \& Res&0.0516 &\textbf{0.0240} &0.0688 &0.0373  &0.0495 &0.0794\\
    MSE w/o Res& 0.0334&\textbf{0.0277} &0.0665 &0.0279 &0.0420 &0.0768 \\
    Huber w/o Res&0.0524 &\textbf{0.0256} &0.0681 &0.0387 &0.0508 &0.0792 \\
    \midrule
    Average&0.0444 &\textbf{0.0255} &0.0713 &0.0382 &0.0469 &0.0821 \\
    \bottomrule
  \end{tabular}
  \label{tab:ab2}
\end{table}
\begin{table}[t]
   \caption{{Ablation study on the hyperparameters. Except for the ablation group, other hyperparameters are optimized settings.}}
  \centering
  \scriptsize
    \setlength{\tabcolsep}{2.5pt}
  \begin{tabular}{l|cccccccc}
    \toprule
    Batch Size &\multicolumn{2}{c}{32}&\multicolumn{2}{c}{64}&\multicolumn{2}{c}{\textbf{128}}&\multicolumn{2}{c}{256}\\
        \midrule
    RMSE &\multicolumn{2}{c}{0.0227}&\multicolumn{2}{c}{0.0231}&\multicolumn{2}{c}{\textbf{0.0225}}&\multicolumn{2}{c}{0.0234}\\
    \midrule
    
    \midrule
    Mis. Rate $p_m$&0.1&\textbf{0.2}&0.3&0.4&0.5&0.6&0.7&0.8\\
        \midrule
    RMSE &0.0226&\textbf{0.0225}&0.0236&0.0239&0.0244&0.0254&0.0257&0.0299\\
    \midrule
    
    \midrule
    Hint Rate $p_h$&0.2&0.3&0.4&0.5&0.6&0.7&0.8&\textbf{0.9}\\
        \midrule
    RMSE &0.0245&0.0244&0.0242&0.0241&0.0234&0.0231&0.0229&\textbf{0.0225}\\
    \midrule

    \midrule
    Huber Thresh. $\delta$&0.2&0.3&0.4&0.5&\textbf{0.6}&0.7&0.8&0.9\\
     \midrule
    RMSE &0.0236&0.0233&0.0231&0.0230&\textbf{0.0225}&0.0226&0.0227&0.0232\\
    \midrule
        
    \midrule
    Loss Weight $\alpha$&2&4&6&8&\textbf{10}&12&14&16\\
     \midrule
    RMSE &0.0248&0.0243&0.0240&0.0238&\textbf{0.0225}&0.0229&0.0231&0.0233\\
    \midrule
    
    \midrule
    Reg. Weight $\lambda$&\multicolumn{2}{c}{0.1}&\multicolumn{2}{c}{0.01}&\multicolumn{2}{c}{\textbf{0.001}}&\multicolumn{2}{c}{0.0001}\\
     \midrule
    RMSE &\multicolumn{2}{c}{0.0239}&\multicolumn{2}{c}{0.0244}&\multicolumn{2}{c}{\textbf{0.0225}}&\multicolumn{2}{c}{0.0229}\\
    \bottomrule
  \end{tabular}
  \label{tab:ab3}
\end{table}
\begin{table}[t]
  \caption{{Ablation study on the rotation strategy.}}
  \centering
  \footnotesize
  \begin{tabular}{l|cc}
    \toprule
    Method& \textbf{SDR-GAIN}&w/o Rotation \\
    \midrule
    RMSE&\textbf{0.0225}&0.0233\\
    \bottomrule
  \end{tabular}
  \label{tab:rot}
\end{table}
\begin{table}[t]
  \caption{{Inference time on different GPUs.}}
  \centering
  \footnotesize
  \begin{tabular}{l|cccccc}
    \toprule
    GPU&3060Ti&T4&L4&A100\\
    \midrule
    Time/$10^{-4}$s&4.580&3.173&3.361&3.337\\
    \bottomrule
  \end{tabular}
  \label{tab:gpus}
\end{table}



\noindent\textbf{{Loss functions and model structures.}} As shown in Table~\ref{tab:ab1}, to explore the optimal loss functions and model structures for training a single generator under different data conditions, we conduct ablation analysis on the performance of models with different loss functions and structures under separation and dimensionality reduction strategies. In this table, ``Head'' and ``Body'' denote training data collected separately from the head and torso, while ``Both'' refers to the full-body training data without separation. The terms ``HeadD,'' ``BodyD,'' and ``BothD'' correspond to the respective training data after dimensionality reduction. Specifically, ``HeadD'' (BodyD/BothD) comprises the dimensionality reduced x- and y-axis coordinates, denoted as ``HeadX'' (BodyX/BothX) and ``HeadY'' (BodyY/BothY). Since the normalized scales of each dimension of data after separation differ, corss-group comparisons are unfair. Therefore, we analyzed the performance of models data types to determine the most suitable model for each data type. The results suggest that the Huber loss function and residual structures generally improve model performance. However, for head-related data in the SDR-GAIN group, better results are achieved without residual structures. Besides, for body data in the w/o DR group, BothX and BothY data in the w/o S group, and data in the w/o S \& DR group, MSE loss yields superior performance. These findings highlight the importance of aligning the choice of loss function and model structure with the specific characteristics of the data.

\noindent\textbf{{Separation, dimensionality reduction 
 and multi-generator strategies.}} As shown in Table~\ref{tab:ab2}, we perform ablation analysis on the effects of separation, dimensionality reduction, and multi-generator strategies under consistent data conditions. In the SDR-GAIN group, “Dev” indicates the use of four generators to produce ``HeadX,'' ``HeadY,'' ``BodyX,'' and ``BodyY,'' separately, while ``No Dev'' uses two generators to generate ``HeadD'' and ``BodyD''. Similarly, in the w/o S group, ``Dev'' refers to two generators for ``BothX'' and ``BothY'', whereas ``No Dev'' uses a single generator for ``BothD''. For the w/o DR group, which omits dimensionality reduction, two generators are employed for the ``Head'' and ``Body''. For the w/o S \& DR group, neither separation nor dimensionality reduction is applied, hence one generator is used to produce the ``Both''. 
 
 However, the multi-generator strategy requires separate normalization for head and body data, which differ in scale. To unify the normalization process, all data must first be converted to real-world coordinates before applying a common normalization scale. While this results in lower RMSE values, it does not affect the validity of performance comparisons. The results demonstrate that the No Dev group achieves superior performance under the SDR-GAIN setting. Therefore, to maximize the effectiveness of SDR-GAIN, it is recommended to train two separate generators for HeadD and BodyD. Both separation and dimensionality reduction contribute to performance gains, with separation having a more pronounced effect. Given the necessity of training two generators, their optimal model structures should be determined based on the analysis in Table~\ref{tab:ab1}. Specifically, for HeadD, we recommend using the Huber loss function without residual structures; for BodyD, the Huber loss function combined with residual structures yields the best results. This configuration should be considered the standard practice when applying the SDR-GAIN method.

\noindent{\textbf{Hyperparameters.} As shown in Table~\ref{tab:ab3}, we conduct an ablation study on hyperparameters and adopt the optimal settings indicated by the results for model training.}

\noindent{\textbf{Rotation.} As shown in Table~\ref{tab:rot}, we conduct an ablation study on the rotation strategy. The results demonstrate that rotating human poses to a consistent angle facilitates more effective learning of pose features by the model.}

 \noindent{\textbf{Inference time on different GPUs.} As shown in Table~\ref{tab:gpus}, we report the inference time of SDR-GAIN on different GPUs, with CPU-related computations measured on an i7-12700F. While more powerful GPUs generally offer faster inference, both L4 and A100 were slightly slower than T4 despite their higher theoretical performance. This is likely due to increased overhead from drivers, scheduling, and initialization, which outweighs the speed gains for very lightweight models like SDR-GAIN.}

\subsection{Discussion}
\begin{table}[t]
  \caption{Running time of pedestrian pose estimation tasks and relative proportion of SDR-GAIN’s time cost as an integrated post-processing module.}
  \centering
  \footnotesize
  \setlength{\tabcolsep}{3pt}
  \begin{tabular}{l|cccc}
    \toprule
    \multirow{2}{*}{Metrcis} & \multicolumn{4}{c}{Pedestrian pose estimation tasks}\\ \cmidrule(l){2-5}
    &PEN-ALFNet-1s&PEN-ALFNet-2s&PEN-CSP&CamLoc \\
    \midrule
    Time/s&0.07&0.08&0.10&0.023\\
    Proportion/\%&0.57&0.50&0.40&1.74\\
    \bottomrule
  \end{tabular}
  \label{tab:dis}
\end{table}

\noindent\textbf{Real-time performance as an integrated post-processing module.} As shown in Table~\ref{tab:dis}, we analyze the proportion of inference time added by SDR-GAIN when integrated as a post-processing module into various pedestrian pose estimation pipelines. The results demonstrate that SDR-GAIN introduces only a minimal increase in inference time. Therefore, incorporating SDR-GAIN does not significantly compromise the real-time performance of the original system, highlighting its practical applicability in time-sensitive scenarios.

\noindent{\textbf{Limitations.} The limited scale of complete pedestrian pose data available for GAN training constrain the performance of SDR-GAIN. As generative models generally require large-scale datasets to achieve optimal performance, utilizing larger or domain-specific datasets could further enhance SDR-GAIN's effectiveness. Additionally, in adversarial learning, the presence of a min-max game often leads to convergence difficulties, such as vanishing gradients, mode collapse, and oscillations in the parameter space, making it challenging to reach a stable equilibrium. A potential solution is to stabilize training through regularization techniques~\cite{WGAN,fu2023brainvis,gu2023orsi,fu2023sgcn,zheng2022boosting}.}

\section{Conclusion}
{This paper presents SDR-GAIN, a novel real-time framework for completing occluded pedestrian poses in autonomous driving scenarios. Leveraging a self-supervised generative adversarial learning approach, SDR-GAIN effectively reconstructs missing keypoints caused by occlusions. The framework integrates lightweight residual-structured generators with pose standardization techniques to enhance feature representation. Experiments on the COCO and JAAD datasets show that SDR-GAIN outperforms both traditional machine learning methods and Transformer-based models, achieving a 47.4\% reduction in RMSE while maintaining microsecond-level inference times. Ablation studies further confirm the effectiveness of components such as the multi-generator strategy and pose standardization in improving accuracy. Despite limitations due to the small dataset size and the training instability inherent in adversarial learning, SDR-GAIN offers a strong balance between accuracy and efficiency, making it a promising solution for enhancing occluded pedestrian detection in autonomous driving systems.}

\bibliographystyle{IEEEtran}
\bibliography{references}

\end{document}